\documentclass{article}

\usepackage{arxiv}

\usepackage[utf8]{inputenc} %
\usepackage[T1]{fontenc}    %
\usepackage{hyperref}       %
\usepackage{url}            %
\usepackage{booktabs}       %
\usepackage{amsfonts}       %
\usepackage{nicefrac}       %
\usepackage{microtype}      %

\usepackage{amsmath,amsfonts,bm}

\def\eqref#1{equation~\ref{#1}}

\def\1{\bm{1}}

\def\vmu{{\bm{\mu}}}

\def\vdelta{{\bm{\delta}}}
\def\va{{\bm{a}}}

\def\vq{{\bm{q}}}

\def\vw{{\bm{w}}}

\def\vz{{\bm{z}}}

\DeclareMathAlphabet{\mathsfit}{\encodingdefault}{\sfdefault}{m}{sl}
\SetMathAlphabet{\mathsfit}{bold}{\encodingdefault}{\sfdefault}{bx}{n}

\def\gA{{\mathcal{A}}}

\def\gD{{\mathcal{D}}}

\def\gX{{\mathcal{X}}}
\def\gY{{\mathcal{Y}}}
\def\gZ{{\mathcal{Z}}}

\def\sH{{\mathbb{H}}}

\def\sP{{\mathbb{P}}}

\def\sR{{\mathbb{R}}}

\usepackage{cleveref}       %
\usepackage{lipsum}         %
\usepackage{graphicx}
\usepackage{natbib}
\usepackage{doi}

\usepackage{amsthm}
\usepackage{natbib}
\usepackage{xcolor}
\usepackage{subcaption}
\usepackage{graphicx}
\usepackage{listings}
\usepackage{algorithm}
\usepackage{algorithmic}
\usepackage{wrapfig}
\newcommand{\mypar}[1]{\vspace{-2mm}\paragraph{#1}}

\title{On the Effects of Adversarial Perturbations on Distribution Robustness}

\date{}

\newif\ifuniqueAffiliation

\ifuniqueAffiliation %
\else
\usepackage{authblk}

\author[]{%
    \hspace{1mm}Yipei Wang%
    \thanks{Equal Contribution}%
}
\author[]{%
    \hspace{1mm}Zhaoying Pan$^*$%
}
\author[]{%
    \hspace{1mm}Xiaoqian Wang%
    \thanks{Corresponding author.}%
}
\affil[1]{Elmore Family School of Electrical and Computer Engineering, Purdue University}

\fi

\begin{document}
\maketitle

\vspace{-5mm}
\begin{abstract}

    Adversarial robustness refers to a model's ability to resist perturbation of inputs, while distribution robustness evaluates the performance of the model under data shifts. Although both aim to ensure reliable performance, prior work has revealed a tradeoff in distribution and adversarial robustness. Specifically, adversarial training might increase reliance on spurious features, which can harm distribution robustness, especially the performance on some underrepresented subgroups. We present a theoretical analysis of adversarial and distribution robustness that provides a tractable surrogate for per-step adversarial training by studying models trained on perturbed data.
    In addition to the tradeoff, our work further identified a nuanced phenomenon that $\ell_\infty$ perturbations on data with moderate bias can yield an increase in distribution robustness. Moreover, the gain in distribution robustness remains on highly skewed data when simplicity bias induces reliance on the core feature, characterized as greater feature separability. Our theoretical analysis extends the understanding of the tradeoff by highlighting the interplay of the tradeoff and the feature separability. 
    Despite the tradeoff that persists in many cases, overlooking the role of feature separability may lead to misleading conclusions about robustness.
    
\end{abstract}

\section{Introduction}
    Despite machine learning having achieved great success, its application in the real world can face many challenges, such as input noise~\citep{gupta2019dealing} or data shifts~\citep{quinonero2022dataset}. The well-established datasets are always curated with clean data and labels. At the same time, the real-world input can be accompanied by various image corruptions, such as noise, blur, or changes in brightness~\citep{hendrycks2018benchmarking}. What's worse, the models can be sensitive to small adversarial perturbations~\citep{szegedy2013intriguing} that are generated according to the trained models on intention, which results in a minimal change in the image appearance but significantly misleads the model prediction. Therefore, such model vulnerabilities call for effective defense strategies~\cite{goodfellow2014explaining, madry2017towards} to improve adversarial robustness, commonly achieved by incorporating adversarial perturbations in the training data.

\begin{figure*}[t!]
    \centering
    \includegraphics[width=0.99\linewidth]{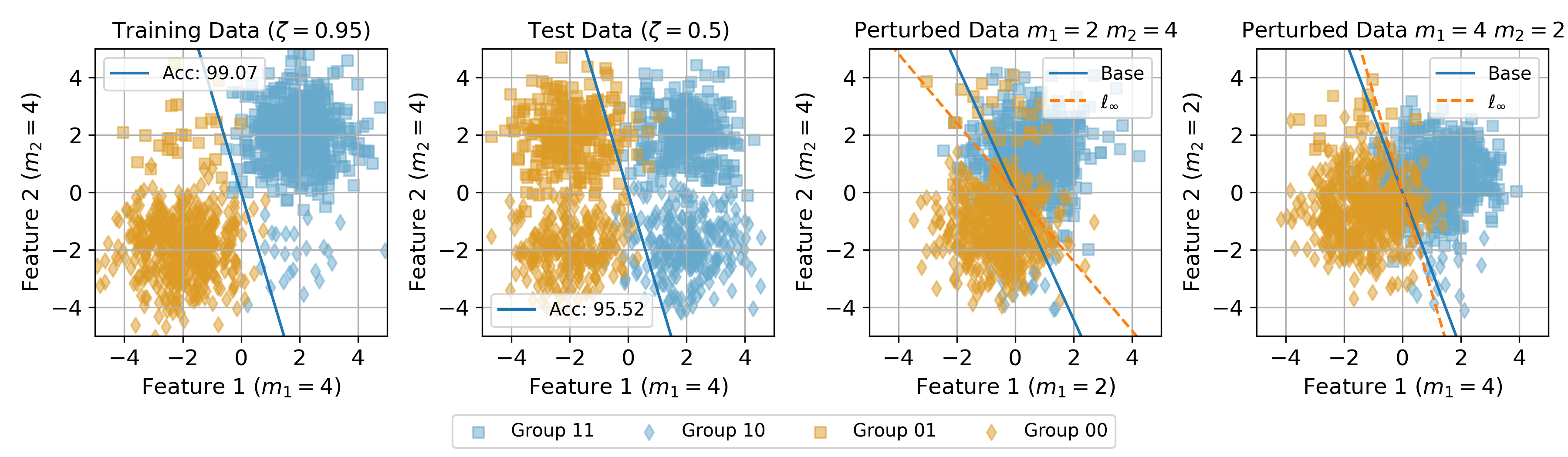}
    \caption{Examples of Distribution Robustness and Adversarial Perturbations. Groups are denoted by their two binary attributes. 
    The left subplots show distribution robustness when training data contains spurious correlation ($\zeta=0.95$; 95\% majority) and test data is balanced, leading to an accuracy drop under shift. The right subplots show training data perturbed with $\ell_\infty$ at varying feature separabilities and a large attack budget of 1 to highlight the effects. When the spurious feature is easier to classify, as $m_1 < m_2$, the classifier increasingly relies on the spurious feature (feature 2); conversely, $m_1 > m_2$ increases the reliance on the non-spurious feature.
    }
    \label{fig:teaser}
    \vspace{-3mm}
\end{figure*}

    In addition to adversarial robustness, robustness to data shift~\citep{quinonero2022dataset, koh2021wilds} is another major concern in real-world applications. It is common to assume that the training and test data are independent and identically distributed (i.i.d.), but it might fail in real-world applications. This disparity between the training and test data, i.e., distribution shifts~\citep{koh2021wilds}, could likely yield a performance drop on the test data or subgroups that are underrepresented in training data. Ideally, the model is expected to learn the core feature to achieve distribution robustness, by making correct predictions even in the presence of distribution shifts between training and testing data~\citep{arjovsky2019invariant}. However, spurious correlation~\citep{sagawa2019distributionally, geirhos2020shortcut}, where the core feature correlates with other features, poses significant challenges in distribution robustness. For data with spurious correlation, there exist context-dependent cues~\citep{beery2018recognition, geirhos2020shortcut, xiao2020noise} that show high correlation with the groundtruth labels in the training data but are not necessarily required for the groundtruth, and do not even hold in some subgroups.
    A canonical example, Waterbirds~\citep{sagawa2019distributionally}, is a dataset for bird species classification of water or land birds, where the training data shows a high correlation between the bird species and backgrounds. However, the background does not guarantee that the bird species and models perform poorly on those groups without this correlation.

    While both adversarial and distribution robustness matter to ensure reliable applications, they are often discussed separately. 
    \cite{moayeri2022explicit} firstly studied the interaction of adversarial and natural distribution robustness, and formulated it as an explicit tradeoff. In particular, adversarial training, despite being considered to improve adversarial robustness effectively~\cite{goodfellow2014explaining, madry2017towards}, actually leads to a decrease in distribution robustness. Specifically, this work showed that this tradeoff happens with $\ell_1, \ell_2$ adversarial training, along with $\ell_\infty$ conditioned on a larger scale for the spurious feature.

    However, the conditions underlying this tradeoff remain unclear, as the loss-based analysis for adversarial training in \citep{moayeri2022explicit} does not fully characterize model performance. To address this gap, we study the effects of adversarial perturbations with a two-stage framework that enables a more precise examination of the interaction between adversarial robustness and distribution robustness. Specifically, we first fit a classifier, then generate adversarial perturbations using its gradients and fit a second classifier on the perturbed data. Motivated by the feature-scale conditions studied in \citep{moayeri2022explicit}, we aim to analyze these roles in greater depth. Building on \cite{wang2024effect}, which hypothesized feature separability as a key factor in spurious correlation modeling and analyzed its impact on accuracy, we adopt a similar perspective in our setting. By examining how the accuracy of the perturbed-data classifier shifts from the clean-data classifier, we investigate the effects and patterns of adversarial perturbations. Following \citep{moayeri2022explicit}, we evaluate distribution robustness by considering skewed training data with spurious correlations, where performance drop happens when the test distribution shifts, for example, to balanced test data, as shown in the two left subplots of \cref{fig:teaser}.
    
    Our theoretical results reveal that both the feature separability and the spurious correlation severity play a vital role in the effects of adversarial perturbations on distribution robustness. We show that, beyond the conventional tradeoff, adversarial perturbations can improve distribution robustness in specific regimes.
    In particular, for moderately biased data where spurious features are correlated with labels but do not fully dominate, $\ell_\infty$ perturbations mitigate reliance on spurious features across a broad range of feature separability. This improvement arises even in an extreme spurious correlation when the core features are easy to learn.

    We validate these theoretical observations using synthetic Gaussian data constructed to match our analytical setting. The experiments are first conducted under the proposed two-stage framework and the results closely match our theoretical insights. Moreover, we conduct the experiments with a standard adversarial training where perturbations are applied per training step, and the results confirm that our two-stage framework serves as an effective and tractable surrogate to study the effects of adversarial training via perturbations.

    \begin{itemize}
        \item We present a theoretical analysis, under a simplified framework, to characterize how adversarial perturbations influences distribution robustness.
        \item We show that feature separability and the severity of spurious correlation are key factors to determine the tradeoff between adversarial and distribution robustness. Moreover, we identified more nuanced patterns of their interplay beyond the tradeoff.
        \item Our experiments on synthesized data validate our theoretical insights and also show that our simplified framework generalizes the insights of adversarial perturbations to the standard setup of adversarial training.
    \end{itemize}

\section{Related Work}

    \paragraph{Spurious correlation in distribution robustness.} 
    The distribution robustness~\cite{koh2021wilds} refers to the capacity of a model to maintain strong performance even when the test distribution shifts.
    Spurious correlation refers to the invariant features correlated with other features in the training data, but it might not hold in test data, thus resulting in a performance drop and hurting the distribution robustness~\cite{arjovsky2019invariant, sagawa2019distributionally, geirhos2020shortcut, wang2024effect}. Models trained on data with spurious correlations might rely on the non-robust attributes or features, which do not always correspond to the groundtruth labels. For example, recent research has revealed that models make use of background~\cite{beery2018recognition, geirhos2020shortcut, xiao2020noise, lu2025think}, texture~\cite{geirhos2018imagenet}, or artifacts~\cite{codella2018skin}, and perform poorly on test data, especially those subgroups without these cues. 
    Some studies~\cite{arjovsky2019invariant, jang2024adversarial, lu2024neural, lu2024debiasing, wang2024effect, lu2025mitigating} have proposed mitigating spurious correlation to increase the distribution robustness via encouraging the models to focus more on core features rather than others.

    \mypar{Adversarial robustness and its unexpected outcome.}
    Prior research~\cite{szegedy2013intriguing, goodfellow2014explaining} has revealed the intriguing vulnerability of models under adversarial perturbations, where the difference between the perturbed and original images is negligible to the naked eye but significantly misleads the model prediction. 
    The gradients with respect to the images, obtained from well-trained classifiers, are often utilized as efficient directions for perturbations~\cite{goodfellow2014explaining, madry2017towards}. To avoid severe image corruption, an attack budget is used to ensure the small scale of the perturbation, and the perturbations are categorized as $\ell_2$ or $\ell_\infty$ according to the constraints on the norms of the gradients~\cite{szegedy2013intriguing, goodfellow2014explaining, carlini2017towards}. Fast gradient sign method (FGSM)~\cite{goodfellow2014explaining} is a simple but effective method to generate adversarial examples under $\ell_\infty$ constraints. 
    \cite{ilyas2019adversarial} suggested that the vulnerability to adversarial perturbations comes from the fact that models rely on non-robust features.
    Applying adversarial perturbation during model training has been proposed as an effective strategy to encourage models to favor robust features and improve adversarial robustness~\cite{goodfellow2014explaining, kurakin2016adversarial, madry2017towards}. 
    However, several works have discussed the negative outcome as costs, including hurting standard accuracy~\cite{tsipras2018robustness, raghunathan2019adversarial, zhang2019theoretically}, fairness~\cite{xu2021robust, ma2022tradeoff}, distribution robustness~\cite{moayeri2022explicit}, or robustness against simple attacks~\cite{duan2023inequality}.
    Notably, our work follows the conclusion drawn in \cite{moayeri2022explicit}, which found that adversarial training can compromise the distributional robustness. Our work further examined the conclusion in a regime of smaller attack budgets and identified counterexamples and their conditions for a more comprehensive understanding of the interaction between adversarial perturbations and distributional robustness.

\section{Distribution Robustness under Adversarial Perturbation}

\subsection{Overview}
    We aim to study the effects of adversarial perturbations on the models' distribution robustness, particularly in the presence of spurious correlations. Following the setup in \cite{moayeri2022explicit, wang2024effect}, we consider the training distribution with spurious correlation, in which the majority of data exhibit a correlation between target labels and some noncausal attributes, while a minority violates the correlation. We categorize the attributes into {\bf invariant} and {\bf spurious} attributes. The invariant attributes capture the core signal that consistently determines the target labels across data distributions, whereas the spurious attributes are correlated with the label only within certain groups and may conflict with the labels in others. 
    
    Models trained under such skewed data often rely on the spurious attributes, resulting in degraded performance on data where the correlation does not hold. Although this degradation may have limited impacts on the overall training loss or accuracy since these groups are typically a minority, it becomes substantial when the test data distribution is shifted from the training distribution, and these minority groups in training data occupy a larger proportion of the test data. 
    The distribution robustness is therefore evaluated by the dependence on the invariant features or the performance under the shifted test data. In our analysis, we consider a balanced test distribution over groups, which corresponds to the adjusted accuracy~\cite{yang2023change}, defined as the mean of per-group accuracies. Our theoretical analysis builds on the data modeling and clean-data Bayes-optimal classifier analysis introduced in \cite{wang2024effect}. We briefly summarize these components for completeness before extending the framework to adversarial perturbations.

    On the other hand, adversarial training, as an effective measure to improve the adversarial robustness of models, applies perturbation to training data during each iteration. Direct theoretical analysis of the adversarial training is challenging due to its bi-level optimization structure. To enable tractable analysis, we propose a two-stage proxy scheme: a classifier is first trained on clean data and used to generate adversarially perturbed samples, after which a second classifier is trained on the perturbed data and evaluated for distribution robustness. This simplified setup allows us to isolate and examine the effects of adversarial perturbations on the model training and performance. The subsequent sections show the interaction between the adversarial and distribution robustness via analyzing whether the adversarial perturbations amplify the reliance on spurious features.

\subsection{Data Modeling}\label{sec:data-modeling}
    Following the conventional setup of the study of spuriousness \citep{nagarajan2020understanding,sagawa2020investigation,yao2022improving,idrissi2022simple,ming2022impact,liu2022self,wang2024effect}, we establish our analysis in the context of binary classification over a dataset $\gD = \gX\times\gY$ with input images as $\gX$ and labels as $Y \in \gY = \{-1, 1\}$ with $Y \sim \text{Uniform}\{-1, 1\}$.
    We assume there are $N$ binary attributes in the images, denoted as $(A_1, \cdots, A_N) \in \gA$, where each attribute $A_n \sim \text{Uniform}\{-1, 1\}$. Without loss of generality, such as by reindexing attributes, let $A_1$ be the invariant attribute that is consistent with the label, i.e., $Y \equiv A_1$. In contrast, the remaining attributes may correlate with the invariant one to varying degrees, which we quantify by $\zeta_n = \sP(A_n = y \mid Y = y)$ and further denote it with $\boldsymbol{\zeta} = (\zeta_1, \cdots, \zeta_N)$.

    Following the setup of data modeling with spurious correlation in \citep{wang2024effect}, we consider modeling the features with Gaussian mixtures.
    we abstract the input $x$ by its feature representation $\vz = (z_1, \cdots, z_N) \in \gZ$ where $Z_n \in \sR^{d_n}$, $\gZ \in \sR^d$ and $d=\sum_{n=1}^N d_n$ is the total feature dimensions. Following prior work~\cite{wang2024effect}, we model the features as Gaussian variables conditioned on the values of attributes:
    \[
    \vz_n \mid y \sim \begin{cases}\mathcal{N}\left(y \vmu_n, \Sigma_n\right), & a_n=y \\ \mathcal{N}\left(-y \vmu_n, \Sigma_n\right), & a_n \neq y\end{cases}
    \]
    where $\vmu_n, \Sigma_n$ are the mean and covariance of the Gaussian variables and model the geometric characteristics of the $n$-th feature.
    As the attribute takes binary values, it creates two clusters of Gaussian features. The further the clusters are, the easier it is for the classifier to separate the attribute. We therefore use the Mahalanobis distance $m_n = \vmu_n^T \Sigma_n^{-1} \vmu_n$ to quantize the separability for the $n$-th feature, where a larger $m_n$ indicates an attribute easier to classify. 

    \begin{wrapfigure}{h}{0.4\textwidth}
\vspace{-9mm}
\begin{minipage}[H]{\linewidth}
\begin{algorithm}[H]
\caption{Two-stage proxy for adversarial training (PyTorch-like pseudocode)}
    \label{ag:proxy}
        \definecolor{codeblue}{rgb}{0.25,0.5,0.5}
        \lstset{
          backgroundcolor=\color{white},
          basicstyle=\fontsize{7.2pt}{7.2pt}\ttfamily\selectfont,
          columns=fullflexible,
          breaklines=true,
          captionpos=b,
          commentstyle=\fontsize{7.2pt}{7.2pt}\color{codeblue},
          keywordstyle=\fontsize{7.2pt}{7.2pt},
          escapechar=\&%
        }
\begin{lstlisting}[language=python]
# Inputs:
#   m1, m2: feature separability parameters
#   zeta: spurious correlation in training data
#   eps: perturbation budget
#   norm: threat model ("l_inf" or "l_2")

# Single-step (FGSM-style) adversarial perturbation
def perturb(clf, x, y, eps, norm):
    # gradient of loss w.r.t. input
    grad_x = grad(loss(clf(x), y), x)
    if norm == "l_inf":
        delta = grad_x.sign()
    elif norm == "l_2":
        delta = grad_x / (grad_x.norm(p=2, dim=1, keepdim=True) + 1e-8)
    return x + eps * delta

# Generate training and test data
train_x, train_y = get_data(m1, m2, zeta)
test_x, test_y = get_data(m1, m2, zeta=0.5)

# Stage 1: train classifier on clean data
clf_clean = train(train_x, train_y)

# Generate adversarially perturbed training data
adv_x = perturb(clf_clean, train_x, train_y, eps, norm)

# Stage 2: train classifier on perturbed data
clf_proxy = train(adv_x, train_y)

# Evaluate distribution robustness
acc_proxy = eval(clf_proxy, test_x, test_y)
\end{lstlisting}
\end{algorithm}
\vspace{-2mm}
\begin{algorithm}[H]
\caption{Standard adversarial training (PyTorch-like pseudocode)}
\label{ag:adv}
        \definecolor{codeblue}{rgb}{0.25,0.5,0.5}
        \lstset{
          backgroundcolor=\color{white},
          basicstyle=\fontsize{7.2pt}{7.2pt}\ttfamily\selectfont,
          columns=fullflexible,
          breaklines=true,
          captionpos=b,
          commentstyle=\fontsize{7.2pt}{7.2pt}\color{codeblue},
          keywordstyle=\fontsize{7.2pt}{7.2pt},
          escapechar=\&%
        }
\begin{lstlisting}[language=python]
# Adversarial training with single-step perturbations
def adv_train(x, y, eps, norm):
    clf = init_clf()
    for epoch in range(epochs):
        for x_batch, y_batch in dataloader(x, y):
            adv_x = perturb(clf, x_batch, y_batch, eps, norm)
            loss = loss_func(clf(adv_x), y_batch)
            update(clf, loss)
    return clf

# Train adversarially robust classifier
clf_adv = adv_train(train_x, train_y, eps, norm)

# Evaluate distribution robustness
acc_adv = eval(clf_adv, test_x, test_y)
\end{lstlisting}
\end{algorithm}
\end{minipage}
\vspace{-30mm}
\end{wrapfigure}

    Categorizing the features according to the attribute values $\va$, there are $2^N$ subgroups of features with a proportion of $r_{\va}=\sP(\mathbf{A}=\va)$. We denote the $\vmu_\va = [y\vmu_1, \cdots,  y a_N \vmu_N] = [\vmu_1, \cdots,  a_1 a_N \vmu_N] \in \sR^d$ and $\Sigma = \operatorname{diag}(\Sigma_1, \cdots, \Sigma_N)\in\sH^{d\times d}_+$, to define the characteristics of a subgroup with attributes $\va$.

    To summarize, we use Gaussian mixtures to model the features of data with spurious correlation, incorporating several key factors, including $m_n$ for feature separability and $\boldsymbol{\zeta}$ for spurious correlation severity.

\subsection{Two-stage Proxy for Adversarial Training}
    Adversarial training aims to perturb the training data during training iterations to improve the adversarial robustness of models. However, the on-the-fly perturbations are generated at each training step based on the current model parameters, making the theoretical analysis challenging. Therefore, we adopt a two-stage framework as a proxy to study the effects of adversarial perturbations on model training. Specifically, we first fit a base classifier on clean training data and then generate adversarially perturbed data using this fixed classifier for training another classifier. This proxy simplifies the analysis and isolates the effect of adversarial perturbations on model training.
    We show the pseudo-code of the two-stage proxy and standard adversarial training in \cref{ag:proxy} and \cref{ag:adv} for comparison. While the proxy simplifies the interactions of model and perturbation updating, we discuss its generalizability empirically in the subsequent section.

\subsection{Bayes-Optimal Linear Classifier on Clean Data}
    For completeness, we restate several results from \cite{wang2024effect}, which form the basis of our subsequent analysis.
    We focus on linear classifiers of the form $\hat{y}=\mathrm{sign}(\vw^T \vz)$ and we characterize the Bayes-optimal classifier given the Gaussian feature representation model following the analysis of ~\cite{wang2024effect}. For a linear classifier parameterized by $\vw$, \cite{wang2024effect} derive a closed-form expression for the training-distribution accuracy as a function of the feature statistics $(\vmu, \Sigma)$ and subgroup parameters $\boldsymbol{\zeta}$.
    \paragraph{Lemma 1 (Training accuracy)~\cite{wang2024effect}.} 
    \textit{The accuracy of a linear classifier $\vw$ under the training distribution is given by
    \begin{align}\label{eq:acc}
        Acc(\vw) =& \frac{1}{2} + \sum_{\va\in\{\pm1\}^N}\Big[r_\va\operatorname{erf}\big(\frac{\vmu_\va^T\vw}{\sqrt{2\vw^T\Sigma \vw}}\big)\Big]
    \end{align}
    where $\va$ indexes a subgroup defined by a specific configuration of the $N$ binary attributes, $r_\va$ denotes the proportion of this subgroup and $\vmu_\va$ is the corresponding conditional mean of the features.
    }

    Optimizing \cref{eq:acc} over the training distribution yields the Bayes-optimal classifier, which is proved to be a highly structured solution: 
    \paragraph{Lemma 2 (Bayes-optimal linear classifier)~\cite{wang2024effect}.}
    \textit{The Bayes-optimal linear classifier under the clean training distribution exhibits block-wise colinearity $\vw^{\#} = \eta_n \Sigma_n^{-1} \mu_n$ for some coefficient $\eta_n \in \sR$.}
    
    This property reduces the optimization over vectors to a scalar optimization of the relative weights. Although the modeling and analysis extend to the case of $N$ attributes, we primarily focus on the two-attribute setting for clarity. Since each spurious attribute is assumed to correlate with the invariant one, insights from the two-attribute case generalize accordingly. For two attributes, we simplify the Bayes-optimal classifier as $\vw^\# = [\Sigma_1^{-1}\vmu_1, \tau\Sigma_2^{-1}\vmu_2]$ with $\tau=\eta_2 / \eta_1$. As $A_1$ is the invariant attribute, $\zeta_1=1$ always holds, and we thus simplify the notation of $\zeta:=\zeta_2$ under the two-attribute setting.

    Given the Bayes-optimal linear classifier $\vw^\#$ under the training distribution, we assess distribution robustness by evaluating its performance on shifted test data. For simplicity, we consider the test accuracy for subgroups characterized by attributes $\va$, which is given by the following immediate consequence of Lemma 1. The test accuracy is therefore obtained by a weighted sum of the group accuracy.
    
    \paragraph{Corollary 3 (Group accuracy under clean data)~\cite{wang2024effect}.}
    \textit{For the two-attribute setting, the accuracy of the Bayes-optimal linear classifier $\vw^\#$ on subgroup $\va$ is
    \begin{equation}\label{eq:group-acc}
        Acc_{\va}(\vw^\#) = \frac{1}{2} \Big(1 + \operatorname{erf}{(\frac{m_1 + a_1 a_2 \tau  m_2}{\sqrt{2(m_1 + \tau^2 m_2)}})}\Big)
    \end{equation}
    }

    In summary, \cite{wang2024effect} characterized the Bayes-optimal linear classifier under the clean training distribution and analyzed its test performance to study distribution robustness. Building on this foundation, we next extend the analysis to adversarial perturbations and examine their impact on the distribution robustness of models.

\subsection{Adversarial Perturbation Generation}

    We generate adversarial perturbations using the Bayes-optimal linear classifier $\vw^\#$ derived in the previous subsection. Specifically, an adversarial perturbation is a bounded input shift that moves the example in the direction that maximally decreases the classification margin with direction determined by $\vw^\#$ and scale controlled by the perturbation budget and constraint. Given a perturbation budget $\epsilon > 0$ and an $\ell_p$ constraint, the adversarial perturbation is applied as $\vz \leftarrow \vz - \epsilon y \Delta^*$ with
    \begin{align}
        \Delta^* = \arg \max_{\|\Delta\|_p \leq 1} (\vw^{\#})^{T} \Delta
    \end{align}

    We consider both $\ell_2$ and $\ell_\infty$ constraints for the perturbation and let $\Delta^*$ have unit norm under the corresponding constraint. Writing $\Delta^* = [\vdelta_1, \vdelta_2]$ for the two-feature setting, the optimal perturbation directions are given by:
    \begin{itemize}
        \item $\ell_2$ constraint:
        \[
        \Delta^* = \frac{\vw^\#}{\|\vw^\#\|_2},
        \quad \text{i.e.,} \quad
        \vdelta_i = \frac{\vw_i^\#}{\|\vw^\#\|_2};
        \]
        \item $\ell_\infty$ constraint:
        \[
        \Delta^* = \operatorname{sign}(\vw^\#),
        \quad \text{i.e.,} \quad
        \vdelta_i = \operatorname{sign}(\vw_i^\#).
        \]
    \end{itemize}
    
    To ensure fair comparison between them, we scale the attack budget for $\ell_2$ perturbations with $\epsilon_2 = \sqrt{d} \cdot \epsilon_{\infty}$, where $d$ is the total feature dimension. This follows from the inequality $\| \Delta \|_2 \leq \sqrt{d} \| \Delta \|_\infty$ and the scaling equalizes the overall perturbation magnitude.

    Under this setup, adversarial perturbations modify the training data distribution through the additional parameter $\epsilon \Delta^*$.
    The perturbed training data are therefore characterized by $(\vmu, \Sigma, \zeta, \epsilon \Delta^*)$, which serves as the basis for analyzing the Bayes-optimal classifier and distribution robustness under adversarial perturbations in the next subsection.

\subsection{Bayes-Optimal Linear Classifier on Perturbed Data}
    Based on the perturbed data, we fit and evaluate a new classifier and analyze the performance compared to that on clean data. 
    Similar to \cref{eq:acc}, we derive the training accuracy on the perturbed data with a linear classifier by:
    \paragraph{Lemma 4 (Training accuracy under perturbed data).}
    \textit{Given the perturbed data distribution characterized by $(\vmu, \Sigma, \zeta, \epsilon\Delta^*)$, the training accuracy of a linear classifier $\vw$ is
    \begin{equation}\label{eq:perturbed-acc}
        \begin{aligned}
            Acc(\vw; \epsilon \Delta) =& \frac{1}{2} + \sum_{\va\in\{\pm1\}^N}\Big[r_\va\operatorname{erf}\big(\frac{(\vmu_\va - \epsilon\Delta^*)^T\vw}{\sqrt{2\vw^T\Sigma\vw}}\big)\Big] \\
        \end{aligned}
    \end{equation}
    }

    This expression serves as the objective function for determining the Bayes-optimal linear classifier under perturbed training data. Maximizing \cref{eq:perturbed-acc} yields a structured solution analogous to the clean-data case, and the Bayesian optimal classifier has a generalized colinearity:
    
    \paragraph{Lemma 5 (Colinearity under perturbed data).} 
    \textit{Let $\vw^*$ maximize $\mathrm{Acc}(\vw; \epsilon \Delta^*)$. In the two-attribute setting, $\vw^*$ follows the form
    \[
    \vw^*
    =
    [c_1\Sigma_1^{-1}(\vmu_1 - \epsilon\vdelta_1),\;
     \Sigma_2^{-1}(c_2\vmu_2 - \epsilon c_1 \vdelta_2)],
    \]
    for some $c_1, c_2 \in \mathbb{R}$.
    }
    
    Similarly, defining $c = c_2/c_1$ which captures the relative weighting between the two attributes, we have the following characterization of the Bayes-optimal linear classifier that fits on perturbed data.

    \paragraph{Theorem 6 (Bayes-optimal linear classifier under perturbed data).} 
    \textit{The coefficient $c$ of the Bayes-optimal linear classifier under adversarial perturbations satisfies the equation
    \begin{align}
        c = \tanh(\frac{\log(\zeta) - \log(1-\zeta)}{2}-\phi(c))
    \end{align}
    where
    \[
    \phi(c)
    =
    \frac{(m_1 - 2\epsilon n_1 - c\epsilon n_2 + \epsilon^2 n_3)
          (\tau m_2 - \epsilon n_2)}
         {m_1 + \tau^2 m_2 - 2\epsilon n_1 - 2c\epsilon n_2 + \epsilon^2 n_3},
    \]
    and
    \[
    n_1 = \vmu_1^T \Sigma_1^{-1}\vdelta_1,\quad
    n_2 = \vmu_2^T \Sigma_2^{-1}\vdelta_2,\quad
    n_3 = (\Delta^*)^T \Sigma^{-1}(\Delta^*) = \vdelta_1^T\Sigma_1^{-1}\vdelta_1 + \vdelta_2^T\Sigma_2^{-1}\vdelta_2.
    \]
    }
    
    It shows that adversarial perturbations modify the clean-data fixed-point equation by introducing additional bias terms that depend on the geometry of the perturbation vector $\Delta^*$.
    Having characterized the Bayes-optimal classifier $\vw^*$ learned from training data perturbed by $\epsilon \Delta^*$, we analyze its performance on individual subgroups in the clean test data.

    \paragraph{Corollary 7 (Group accuracy under perturbed data).}
    \textit{For the two-attribute setting, the accuracy of the Bayes-optimal linear classifier $\vw^*$ on subgroup $\va$ is
    \begin{align}
        \begin{aligned}
            & Acc_{\va}(\vw^*) 
            = \frac{1}{2} + \operatorname{erf}\big(\frac{m_1 + a_1 a_2 c m_2 - \epsilon n_1 - \epsilon a_1 a_2 n_2}{\sqrt{2m_1 + c^2 m_2 - 2\epsilon n_1 - 2c\epsilon n_2 + \epsilon^2 n_3}}\big)
        \end{aligned}
    \end{align}
    }
    
    Corollary 7 explicitly shows how adversarial perturbations introduce accuracy shifts with the perturbation-alignment terms $n_1,n_2,n_3$. In contrast to the clean-data case, perturbations might amplify or suppress performance across groups.
    
    \paragraph{Corollary 8 (Consistency with the clean-data analysis).}
    \textit{Setting $\epsilon = 0$ in the derived Bayes-optimal classifier and group accuracy recovers the clean-data Bayes-optimal classifier and group accuracy characterized in Section 3.4.
    }
    This consistency confirms that the perturbed framework is a strict generalization of the clean-data analysis.

    We now assess the distribution robustness shift induced by the perturbed data by comparing the test accuracy gap obtained from the clean and perturbed classifiers, as $Acc_{\va}(\vw^*) - Acc_{\va}(\vw^\#)$. For simplicity, we assume the test data to be balanced, which shifts from the training data with spurious correlation. We formalize this effect by examining the accuracy gap across a range of values for feature separability $m_1, m_2$ and spurious correlation severity $\zeta$, which characterizes how it shifts differently under $\ell_2$ and $\ell_\infty$ constraints.
    
\begin{figure*}[t!]
    \centering
    \includegraphics[width=0.99\linewidth]{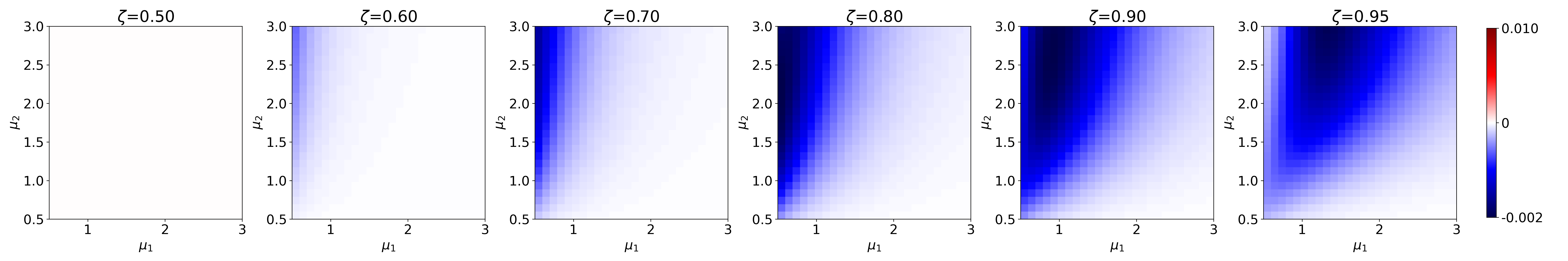}
    \caption{Theoretical accuracy gap between proxy-trained and clean-trained models under $\ell_2$ perturbations.}
    \label{fig:t-acc-norm}
\end{figure*}

\begin{figure*}[t!]
    \centering
    \includegraphics[width=0.99\linewidth]{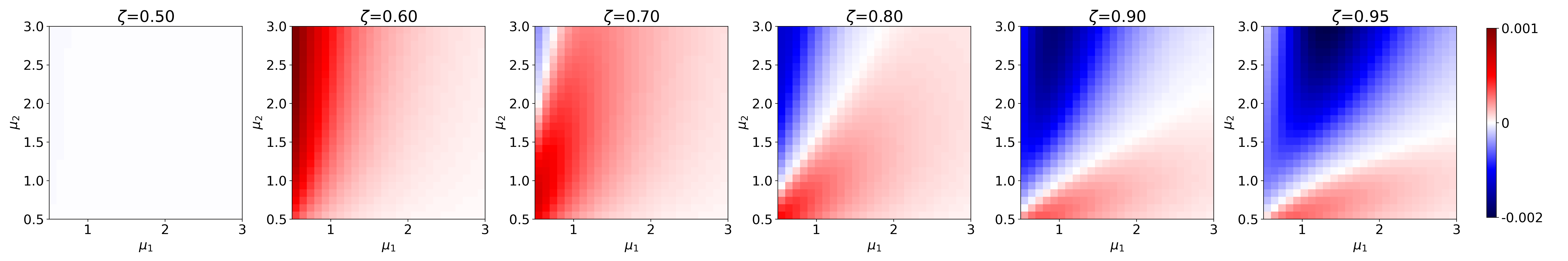}
    \caption{Theoretical accuracy gap between proxy-trained and clean-trained models under $\ell_\infty$ perturbations.}
    \label{fig:t-acc-sign}
\end{figure*}

    \paragraph{Implication (Distribution robustness shift under perturbed data).} 
    Considering $\Sigma=I$, $\mu_1, \mu_2 \in [0.5, 3]$ with $\epsilon=0.01$, we show the visualization of the test accuracy gaps obtained by solving classifiers on data perturbed under $\ell_2$ or $\ell_\infty$ constraints with Corollaries 3 and 7.
    The results are shown in \cref{fig:t-acc-norm} and \cref{fig:t-acc-sign}, where an area of red indicates the conditions when the tradeoff breaks and the perturbations yield an increase in distribution robustness. When $m_2 > m_1$ with severe spurious correlation, both $\ell_2$ and $\ell_\infty$ constraints can reduce distribution robustness; otherwise, the $\ell_\infty$ may increase robustness under certain conditions, including weak spurious correlation with smaller $\epsilon$ or $m_1 > m_2$ under strong spurious correlation.

\section{Experiments}

    Our analysis studies how the robustness varies as three key factors, spurious correlation severity $\zeta$, feature separability $m_1, m_2$, in the data vary; however, in real data, although the spurious correlation strength is controllable, the feature separability is difficult to evaluate or adjust. Therefore, we create synthetic data according to our theoretical setup to validate the insights. We construct the Gaussian features as described in \cref{sec:data-modeling} with the total dimensions of 2 to compare with theoretical results in \cref{fig:t-acc-norm} and \cref{fig:t-acc-sign}. Regarding the model, we use a linear classifier following both the two-stage proxy and adversarial training, and train it for 10 epochs for both the clean and perturbed classifiers in the two-stage proxy, and 100 epochs for adversarial training. We use the L-BFGS optimizer with a learning rate of 1 for both setups, which is consistent with the default solver used in scikit-learn’s logistic regression and enables faster and smoother convergence on small-scale datasets.

\begin{figure*}[t!]
    \centering
    \includegraphics[width=0.99\linewidth]{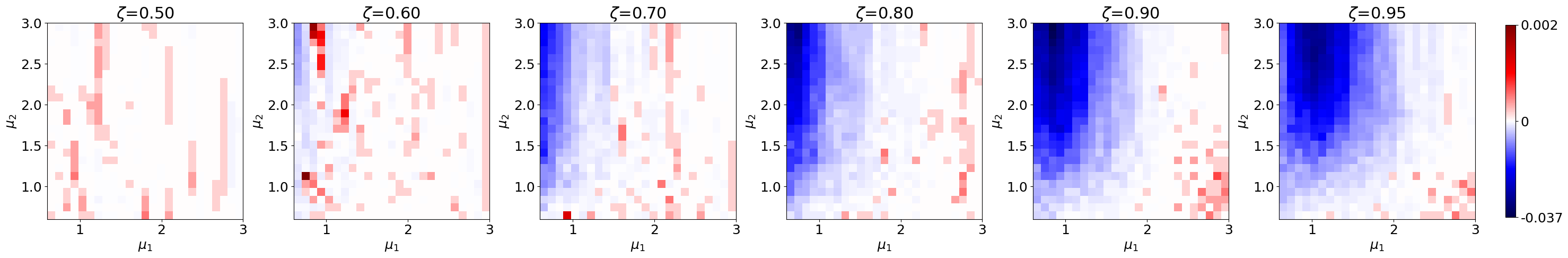}
    \caption{Empirical accuracy gap between proxy-trained and clean-trained models under $\ell_2$ adversarial perturbations.}
    \label{fig:norm}
\end{figure*}

\begin{figure*}[t!]
    \centering
    \includegraphics[width=0.99\linewidth]{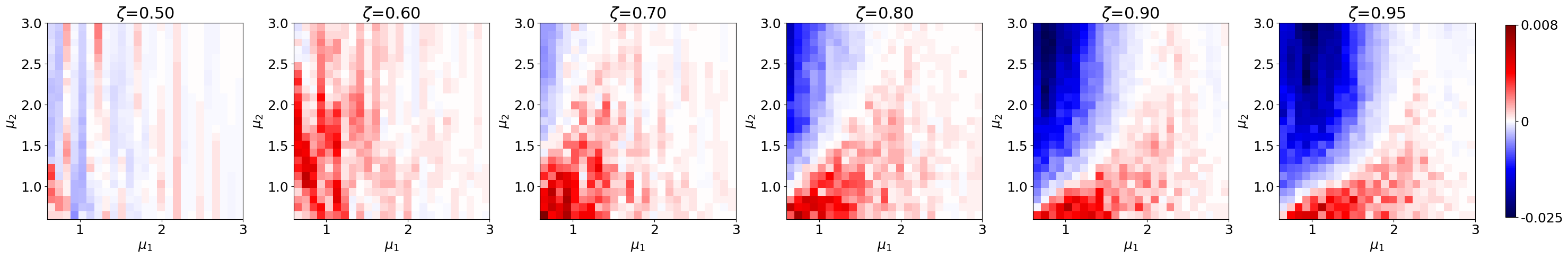}
    \caption{Empirical accuracy gap between proxy-trained and clean-trained models under $\ell_\infty$ adversarial perturbations.}
    \label{fig:sign}
\end{figure*}

\begin{figure*}[t!]
    \centering
    \includegraphics[width=0.99\linewidth]{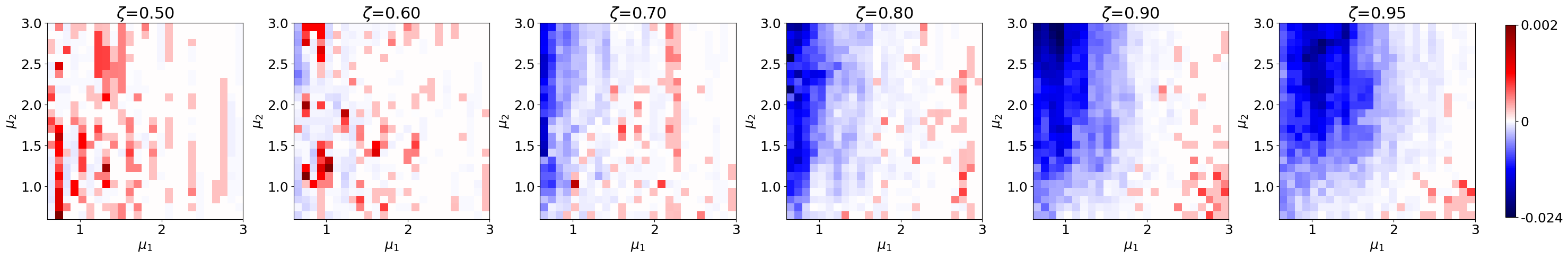}
    \caption{Empirical accuracy gap between adversarially trained and clean-trained models under $\ell_2$ adversarial perturbations.}
    \label{fig:norm-adv}
\end{figure*}

\begin{figure*}[t!]
    \centering
    \includegraphics[width=0.99\linewidth]{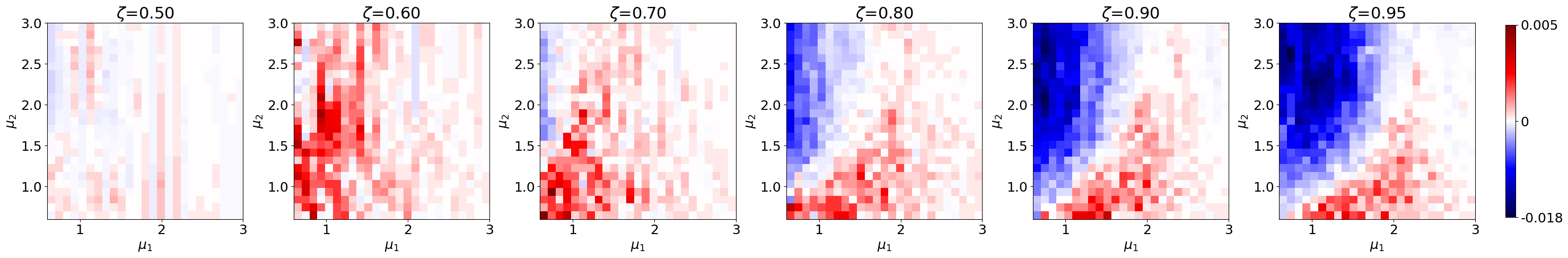}
    \caption{Empirical accuracy gap between adversarially trained and clean-trained models under $\ell_\infty$ perturbations.}
    \label{fig:sign-adv}
\end{figure*}

    Following the two-stage scheme, we conducted experiments to evaluate the accuracy disparity as a result of adversarial perturbations on distribution robustness. We apply the classifier on perturbed and clean data and measure the accuracy disparity, and the results are visualized in \cref{fig:norm} and \cref{fig:sign}. Despite the presence of some red blocks in the $\ell_2$ results, these patterns lack local consistency and do not persist in the nearby configurations, suggesting that they are likely due to empirical noise.
    Overall, the results show consistency with the theoretical results, indicating that our insights reveal a more comprehensive understanding of the effects of adversarial perturbation on distribution robustness. Additionally, we further validate the generalization of the two-stage proxy to the adversarial training that perturbs data in each training step. The results of perturbing per step with the classifier are shown in \cref{fig:norm-adv} and \cref{fig:sign-adv}, matching the theoretical and two-stage results, showing that the insights derived from our simplification with the two-stage scheme can generalize well to the setup of standard adversarial training.

\section{Discussion}
    Our work provides a theoretical analysis of the interplay between adversarial and distribution robustness. We study the distribution robustness under the data modeled with spurious correlation and propose to simplify the adversarial training with a two-stage training scheme. Our analysis shows that feature separability plays a central role in shaping interactions between the adversarial and distribution robustness, beyond the conventional tradeoff. In particular, we demonstrate that $\ell_\infty$ perturbations on moderately biased data can improve distribution robustness, and that this gain can remain in highly skewed settings with a more separable core feature. Empirical results on synthetic data validate our theoretical findings and confirm that the proposed surrogate analysis captures key insights of standard adversarial training.
    These findings provide a principled analysis and clarify when the adversarial–distribution robustness tradeoff holds, highlighting feature separability as a key mediating factor.

    \paragraph{Broader Impact} As prior work proposed the tradeoff between the adversarial and distributional robustness, our work further studied the unexpected effects where the tradeoff breaks, which could provide a more comprehensive understanding for practitioners when applying adversarial perturbations. Furthermore, our work provides a framework to study the $\ell_2$ and $\ell_\infty$ perturbations on data modeled with spurious correlation. Although it's not our focus in this paper, it might be interesting to study further the disparity between $\ell_2$ and $\ell_\infty$ perturbations in future work.

    \paragraph{Limitations.} To enable tractable theoretical analysis, we simplified adversarial training into a two-stage process: first fitting a classifier, then generating perturbations with its gradient to train another classifier. Although our empirical results suggest that this scheme generalizes to adversarial training where perturbations are applied at every step during training, a complete theoretical characterization of standard adversarial training with multiple perturbing steps remains an open problem. Moreover, as the feature separability plays a critical role in our analysis, it remains challenging to quantify the feature separability in the complex setting of real data and adversarial training. Thus our experiments are primarily conducted with the synthetic data, and we leave the examination of complex data for future work.

\bibliographystyle{unsrt}
\bibliography{references}  %

@article{moayeri2022explicit,
  title={Explicit tradeoffs between adversarial and natural distributional robustness},
  author={Moayeri, Mazda and Banihashem, Kiarash and Feizi, Soheil},
  journal={Advances in Neural Information Processing Systems},
  volume={35},
  pages={38761--38774},
  year={2022}
}

@article{gupta2019dealing,
  title={Dealing with noise problem in machine learning data-sets: A systematic review},
  author={Gupta, Shivani and Gupta, Atul},
  journal={Procedia Computer Science},
  volume={161},
  pages={466--474},
  year={2019},
  publisher={Elsevier}
}

@article{yang2023change,
  title={Change is hard: A closer look at subpopulation shift},
  author={Yang, Yuzhe and Zhang, Haoran and Katabi, Dina and Ghassemi, Marzyeh},
  journal={arXiv preprint arXiv:2302.12254},
  year={2023}
}

@book{quinonero2022dataset,
  title={Dataset shift in machine learning},
  author={Qui{\~n}onero-Candela, Joaquin and Sugiyama, Masashi and Schwaighofer, Anton and Lawrence, Neil D},
  year={2022},
  publisher={Mit Press}
}

@inproceedings{xiao2020noise,
  title={Noise or Signal: The Role of Image Backgrounds in Object Recognition},
  author={Xiao, Kai Yuanqing and Engstrom, Logan and Ilyas, Andrew and Madry, Aleksander},
  booktitle={International Conference on Learning Representations},
  year={2020}
}

@article{tsipras2018robustness,
  title={Robustness may be at odds with accuracy},
  author={Tsipras, Dimitris and Santurkar, Shibani and Engstrom, Logan and Turner, Alexander and Madry, Aleksander},
  journal={arXiv preprint arXiv:1805.12152},
  year={2018}
}

@inproceedings{geirhos2018imagenet,
  title={ImageNet-trained CNNs are biased towards texture; increasing shape bias improves accuracy and robustness},
  author={Geirhos, Robert and Rubisch, Patricia and Michaelis, Claudio and Bethge, Matthias and Wichmann, Felix A and Brendel, Wieland},
  booktitle={International conference on learning representations},
  year={2018}
}

@inproceedings{codella2018skin,
  title={Skin lesion analysis toward melanoma detection: A challenge at the 2017 international symposium on biomedical imaging (isbi), hosted by the international skin imaging collaboration (isic)},
  author={Codella, Noel CF and Gutman, David and Celebi, M Emre and Helba, Brian and Marchetti, Michael A and Dusza, Stephen W and Kalloo, Aadi and Liopyris, Konstantinos and Mishra, Nabin and Kittler, Harald and others},
  booktitle={2018 IEEE 15th international symposium on biomedical imaging (ISBI 2018)},
  pages={168--172},
  year={2018},
  organization={IEEE}
}

@article{geirhos2020shortcut,
  title={Shortcut learning in deep neural networks},
  author={Geirhos, Robert and Jacobsen, J{\"o}rn-Henrik and Michaelis, Claudio and Zemel, Richard and Brendel, Wieland and Bethge, Matthias and Wichmann, Felix A},
  journal={Nature Machine Intelligence},
  volume={2},
  number={11},
  pages={665--673},
  year={2020},
  publisher={Nature Publishing Group UK London}
}

@inproceedings{beery2018recognition,
  title={Recognition in terra incognita},
  author={Beery, Sara and Van Horn, Grant and Perona, Pietro},
  booktitle={Proceedings of the European conference on computer vision (ECCV)},
  pages={456--473},
  year={2018}
}

@inproceedings{lu2024neural,
  title={Neural collapse inspired debiased representation learning for min-max fairness},
  author={Lu, Shenyu and Chai, Junyi and Wang, Xiaoqian},
  booktitle={Proceedings of the 30th ACM SIGKDD Conference on Knowledge Discovery and Data Mining},
  pages={2048--2059},
  year={2024}
}

@inproceedings{lu2024debiasing,
  title={Debiasing attention mechanism in transformer without demographics},
  author={Lu, Shenyu and Wang, Yipei and Wang, Xiaoqian},
  year={2024},
  organization={The International Conference on Learning Representations}
}

@inproceedings{lu2025mitigating,
  title={Mitigating Spurious Correlations in Zero-Shot Multimodal Models},
  author={Lu, Shenyu and Chai, Junyi and Wang, Xiaoqian},
  booktitle={The Thirteenth International Conference on Learning Representations},
  year={2025}
}

@inproceedings{jang2024adversarial,
  title={Adversarial fairness network},
  author={Jang, Taeuk and Wang, Xiaoqian and Huang, Heng},
  booktitle={Proceedings of the AAAI Conference on Artificial Intelligence},
  volume={38},
  number={20},
  pages={22159--22166},
  year={2024}
}

@article{arjovsky2019invariant,
  title={Invariant risk minimization},
  author={Arjovsky, Martin and Bottou, L{\'e}on and Gulrajani, Ishaan and Lopez-Paz, David},
  journal={arXiv preprint arXiv:1907.02893},
  year={2019}
}

@inproceedings{koh2021wilds,
  title={Wilds: A benchmark of in-the-wild distribution shifts},
  author={Koh, Pang Wei and Sagawa, Shiori and Marklund, Henrik and Xie, Sang Michael and Zhang, Marvin and Balsubramani, Akshay and Hu, Weihua and Yasunaga, Michihiro and Phillips, Richard Lanas and Gao, Irena and others},
  booktitle={International conference on machine learning},
  pages={5637--5664},
  year={2021},
  organization={PMLR}
}

@article{sagawa2019distributionally,
  title={Distributionally robust neural networks for group shifts: On the importance of regularization for worst-case generalization},
  author={Sagawa, Shiori and Koh, Pang Wei and Hashimoto, Tatsunori B and Liang, Percy},
  journal={arXiv preprint arXiv:1911.08731},
  year={2019}
}

@article{madry2017towards,
  title={Towards deep learning models resistant to adversarial attacks},
  author={Madry, Aleksander and Makelov, Aleksandar and Schmidt, Ludwig and Tsipras, Dimitris and Vladu, Adrian},
  journal={arXiv preprint arXiv:1706.06083},
  year={2017}
}

@article{nagarajan2020understanding,
  title={Understanding the failure modes of out-of-distribution generalization},
  author={Nagarajan, Vaishnavh and Andreassen, Anders and Neyshabur, Behnam},
  journal={arXiv preprint arXiv:2010.15775},
  year={2020}
}

@inproceedings{sagawa2020investigation,
  title={An investigation of why overparameterization exacerbates spurious correlations},
  author={Sagawa, Shiori and Raghunathan, Aditi and Koh, Pang Wei and Liang, Percy},
  booktitle={International Conference on Machine Learning},
  pages={8346--8356},
  year={2020},
  organization={PMLR}
}

@inproceedings{yao2022improving,
  title={Improving out-of-distribution robustness via selective augmentation},
  author={Yao, Huaxiu and Wang, Yu and Li, Sai and Zhang, Linjun and Liang, Weixin and Zou, James and Finn, Chelsea},
  booktitle={International Conference on Machine Learning},
  pages={25407--25437},
  year={2022},
  organization={PMLR}
}

@inproceedings{idrissi2022simple,
  title={Simple data balancing achieves competitive worst-group-accuracy},
  author={Idrissi, Badr Youbi and Arjovsky, Martin and Pezeshki, Mohammad and Lopez-Paz, David},
  booktitle={Conference on Causal Learning and Reasoning},
  pages={336--351},
  year={2022},
  organization={PMLR}
}

@article{szegedy2013intriguing,
  title={Intriguing properties of neural networks},
  author={Szegedy, Christian and Zaremba, Wojciech and Sutskever, Ilya and Bruna, Joan and Erhan, Dumitru and Goodfellow, Ian and Fergus, Rob},
  journal={arXiv preprint arXiv:1312.6199},
  year={2013}
}

@article{goodfellow2014explaining,
  title={Explaining and harnessing adversarial examples},
  author={Goodfellow, Ian J and Shlens, Jonathon and Szegedy, Christian},
  journal={arXiv preprint arXiv:1412.6572},
  year={2014}
}

@inproceedings{lu2025think,
  title={Think Twice: Test-Time Reasoning for Robust CLIP Zero-Shot Classification},
  author={Lu, Shenyu and Pan, Zhaoying and Wang, Xiaoqian},
  booktitle={Proceedings of the IEEE/CVF International Conference on Computer Vision},
  pages={2919--2929},
  year={2025}
}

@inproceedings{carlini2017towards,
  title={Towards evaluating the robustness of neural networks},
  author={Carlini, Nicholas and Wagner, David},
  booktitle={2017 ieee symposium on security and privacy (sp)},
  pages={39--57},
  year={2017},
  organization={Ieee}
}

@article{kurakin2016adversarial,
  title={Adversarial machine learning at scale},
  author={Kurakin, Alexey and Goodfellow, Ian and Bengio, Samy},
  journal={arXiv preprint arXiv:1611.01236},
  year={2016}
}

@inproceedings{
duan2023inequality,
title={Inequality phenomenon in \$l\_\{{\textbackslash}infty\}\$-adversarial training, and its unrealized threats},
author={Ranjie Duan and YueFeng Chen and Yao Zhu and Xiaojun Jia and Rong Zhang and Hui Xue'},
booktitle={The Eleventh International Conference on Learning Representations },
year={2023},
url={https://openreview.net/forum?id=4t9q35BxGr}
}

@inproceedings{zhang2019theoretically,
  title={Theoretically principled trade-off between robustness and accuracy},
  author={Zhang, Hongyang and Yu, Yaodong and Jiao, Jiantao and Xing, Eric and El Ghaoui, Laurent and Jordan, Michael},
  booktitle={International conference on machine learning},
  pages={7472--7482},
  year={2019},
  organization={PMLR}
}

@inproceedings{xu2021robust,
  title={To be robust or to be fair: Towards fairness in adversarial training},
  author={Xu, Han and Liu, Xiaorui and Li, Yaxin and Jain, Anil and Tang, Jiliang},
  booktitle={International conference on machine learning},
  pages={11492--11501},
  year={2021},
  organization={PMLR}
}

@article{ilyas2019adversarial,
  title={Adversarial examples are not bugs, they are features},
  author={Ilyas, Andrew and Santurkar, Shibani and Tsipras, Dimitris and Engstrom, Logan and Tran, Brandon and Madry, Aleksander},
  journal={Advances in neural information processing systems},
  volume={32},
  year={2019}
}

@article{ma2022tradeoff,
  title={On the tradeoff between robustness and fairness},
  author={Ma, Xinsong and Wang, Zekai and Liu, Weiwei},
  journal={Advances in Neural Information Processing Systems},
  volume={35},
  pages={26230--26241},
  year={2022}
}

@inproceedings{ming2022impact,
  title={On the impact of spurious correlation for out-of-distribution detection},
  author={Ming, Yifei and Yin, Hang and Li, Yixuan},
  booktitle={Proceedings of the AAAI Conference on Artificial Intelligence},
  volume={36},
  number={9},
  pages={10051--10059},
  year={2022}
}

@article{raghunathan2019adversarial,
  title={Adversarial training can hurt generalization},
  author={Raghunathan, Aditi and Xie, Sang Michael and Yang, Fanny and Duchi, John C and Liang, Percy},
  journal={arXiv preprint arXiv:1906.06032},
  year={2019}
}

@inproceedings{wang2024effect,
  title={On the Effect of Key Factors in Spurious Correlation: A theoretical Perspective},
  author={Wang, Yipei and Wang, Xiaoqian},
  booktitle={International Conference on Artificial Intelligence and Statistics},
  pages={3745--3753},
  year={2024},
  organization={PMLR}
}

@inproceedings{liu2022self,
  title={Self-supervised Learning is More Robust to Dataset Imbalance},
  author={Liu, Hong and HaoChen, Jeff Z and Gaidon, Adrien and Ma, Tengyu},
  booktitle={International Conference on Learning Representations},
  year={2022}
}

@article{hendrycks2018benchmarking,
  title={Benchmarking neural network robustness to common corruptions and surface variations},
  author={Hendrycks, Dan and Dietterich, Thomas G},
  journal={arXiv preprint arXiv:1807.01697},
  year={2018}
}

\clearpage

\appendix
\section{Appendix}

\subsection{Proof of Lemma 4}

    \paragraph{Lemma 4 (Training accuracy under perturbed data).}
    \textit{Given the perturbed data distribution characterized by $(\vmu, \Sigma, \zeta, \epsilon\Delta^*)$, the training accuracy of a linear classifier $\vw$ is
    \begin{equation}
        \begin{aligned}
            Acc(\vw; \epsilon \Delta^*) =& \frac{1}{2} \Big( 1 + \sum_{\va\in\{\pm1\}^N}\Big[r_\va\operatorname{erf}\big(\frac{(\vmu_\va - \epsilon\Delta^*)^T\vw}{\sqrt{2\vw^T\Sigma\vw}}\big)\Big] \Big) \\
        \end{aligned}
    \end{equation}
    }

    \begin{proof}
        Given $N$ attributes $a_1,\cdots,a_N$, there are totally $2^N$ groups, indexed by values of the binary attributes denoted as $\va$. Let $r_\va$ denote the proportion of this subgroup, and $\vmu_\va = [a_1 \vmu_1, \cdots,  a_N \vmu_N]$ is the corresponding conditional mean of the features for the clean data. The perturbed data, characterized by $(\vmu, \Sigma, \zeta, \epsilon\Delta^*)$, has the conditional mean as $\vmu_\va - \epsilon y \Delta^*$
        The joint distribution of the data conditioned on $y$ can be written as
        \begin{equation}
        \begin{aligned}
        & p(\boldsymbol{z} \mid y=1)=\prod_{n=1}^N p\left(z_n \mid y=1\right) \\
        = & \frac{1}{\sqrt{(2 \pi)^d|\Sigma|}} \sum_{\va \in\{-1,1\}^N}\left\{r_\va \exp \left(-\frac{1}{2}\left(\boldsymbol{z}-\vmu_{\va} + \epsilon \Delta^*\right)^T \Sigma^{-1}\left(\boldsymbol{z}-\vmu_{\va} + \epsilon \Delta^*\right)^T\right)\right\} \\
        & p(\boldsymbol{z} \mid y=-1)=\prod_{n=1}^N p\left(z_n \mid y=-1\right) \\
        = & \frac{1}{\sqrt{(2 \pi)^d|\Sigma|}} \sum_{\va \in\{-1,1\}^N}\left\{r_\va \exp \left(-\frac{1}{2}\left(\boldsymbol{z} + \vmu_{\va} - \epsilon \Delta^*\right)^T \Sigma^{-1}\left(\boldsymbol{z} + \vmu_{\va} - \epsilon \Delta^*\right)^T\right)\right\}
        \end{aligned}
        \end{equation}
    
        Similar to \cite{wang2024effect}, the accuracy can be decomposed as the sum of true positive and true negative rates. Let $\Omega(\vz)=\{\vw \mid \vw^T \vz > 0 \}$ be the half-space of the positive prediction and the true positive rate is
        
        \begin{equation}
        \begin{aligned}
        & \int_{\Omega(\vz)} \frac{1}{\sqrt{(2 \pi)^d|\Sigma|}} \sum_{\va \in\{-1,1\}^N}\left\{r_\va \exp \left(-\frac{1}{2}\left(\vz-\vmu_{\va} + \epsilon \Delta^*\right)^T \Sigma^{-1}\left(\vz-\vmu_{\va} + \epsilon \Delta^*\right)^T\right)\right\} \mathrm{d}\vz \\
        = & \frac{1}{2} \sum_{\va \in\{-1,1\}^N}\left\{r_\va \Big[1 - \operatorname{erf}\Big( -\frac{(\vmu_\va - \epsilon\Delta^*)^T\vw}{\sqrt{2\vw^T\Sigma\vw}}\Big)\Big]\right\} \\
        = & \frac{1}{2} \Big( 1 + \sum_{\va \in\{-1,1\}^N}\Big[r_\va \operatorname{erf} \Big(\frac{(\vmu_\va - \epsilon\Delta^*)^T\vw}{\sqrt{2\vw^T\Sigma\vw}}\Big)\Big] \Big)
        \end{aligned}
        \end{equation}
    
        Similarly, the true negative rate can be written as
        
        \begin{equation}
        \begin{aligned}
        & \int_{\sR^d \backslash \Omega(\vz)} \frac{1}{\sqrt{(2 \pi)^d|\Sigma|}} \sum_{\va \in\{-1,1\}^N}\left\{r_\va \exp \left(-\frac{1}{2}\left(\vz + \vmu_{\va} - \epsilon \Delta^*\right)^T \Sigma^{-1}\left(\vz + \vmu_{\va} - \epsilon \Delta^*\right)^T\right)\right\} \mathrm{d}\vz\\
        = & 1 - \int_{\Omega(\vz)} \frac{1}{\sqrt{(2 \pi)^d|\Sigma|}} \sum_{\va \in\{-1,1\}^N}\left\{r_\va \exp \left(-\frac{1}{2}\left(\vz + \vmu_{\va} - \epsilon \Delta^*\right)^T \Sigma^{-1}\left(\vz + \vmu_{\va} - \epsilon \Delta^*\right)^T\right)\right\} \mathrm{d}\vz\\
        = & \frac{1}{2} \Big( 1 - \sum_{\va \in\{-1,1\}^N}\Big[r_\va \operatorname{erf} \Big(\frac{(-\vmu_\va + \epsilon\Delta^*)^T\vw}{\sqrt{2\vw^T\Sigma\vw}}\Big)\Big] \Big)\\
        = & \frac{1}{2} \Big( 1 + \sum_{\va \in\{-1,1\}^N}\Big[r_\va \operatorname{erf} \Big(\frac{(\vmu_\va - \epsilon\Delta^*)^T\vw}{\sqrt{2\vw^T\Sigma\vw}}\Big)\Big] \Big)
        \end{aligned}
        \end{equation}
    
        Putting them together, we have the accuracy as
        \begin{equation}
            \begin{aligned}
                Acc(\vw; \epsilon \Delta^*) = & \frac{1}{2}\left(\int_{\Omega(\vw)} p(\vz \mid y=1) \mathrm{d} \vz+\int_{\mathbb{R}^d \backslash \Omega(\vw)} p(\vz \mid y=-1) \mathrm{d} \vz\right) \\
                =& \frac{1}{2} \Big( 1 + \sum_{\va\in\{\pm1\}^N}\Big[r_\va\operatorname{erf}\big(\frac{(\vmu_\va - \epsilon\Delta^*)^T\vw}{\sqrt{2\vw^T\Sigma\vw}}\big)\Big] \Big) \\
            \end{aligned}
        \end{equation}
    
    \end{proof}

\subsection{Proof of Lemma 5}
    \paragraph{Lemma 5 (Colinearity under perturbed data).} 
    \textit{Let $\vw^*$ maximize $\mathrm{Acc}(\vw; \epsilon \Delta^*)$. In the two-attribute setting, $\vw^*$ follows the form
    \[
    \vw^*
    =
    [c_1\Sigma_1^{-1}(\vmu_1 - \epsilon\vdelta_1),\;
     \Sigma_2^{-1}(c_2\vmu_2 - \epsilon c_1 \vdelta_2)],
    \]
    for some $c_1, c_2 \in \mathbb{R}$.
    }
    \begin{proof}
        Considering $\vw^*$ maximize $\mathrm{Acc}(\vw; \epsilon \Delta^*)$, we compute the stationary point $\nabla_\vw Acc(\vw; \epsilon \Delta^*)=0$ as
        \begin{equation}
        \begin{aligned}
            0=&\nabla_\vw Acc(\vw; \epsilon \Delta^*)\\
            =& \frac{1}{2} \nabla_\vw \Big( \sum_{\va\in\{\pm1\}^N}\Big[r_\va\operatorname{erf}\big(\frac{(\vmu_\va - \epsilon\Delta^*)^T\vw}{\sqrt{2\vw^T\Sigma\vw}}\big)\Big]\Big) \\
            =&\frac{1}{\sqrt{\pi}} \frac{\Sigma \vw \vw^T-\left(\vw^T \Sigma \vw\right) I}{\left(\vw^T \Sigma \vw\right)^{3 / 2}} \sum_{\va \in\{ \pm 1\}^N}\left\{\gamma_{\va} \exp \left(-\frac{\left((\vmu_\va - \epsilon\Delta^*)^T \vw\right)^2}{2 \vw^T \Sigma \vw}\right) (\vmu_\va - \epsilon\Delta^*)\right\}
        \end{aligned}
        \end{equation}
        which suffices to find $\Sigma \vw \vw^T \vq = \left(\vw^T \Sigma \vw\right) \vq$ with
        \begin{equation}
        \begin{aligned}
        \vq =& \sum_{\va \in\{ \pm 1\}^N}\left\{\gamma_{\va} \exp \left(-\frac{\left((\vmu_\va - \epsilon\Delta^*)^T \vw\right)^2}{2 \vw^T \Sigma \vw}\right) (\vmu_\va - \epsilon\Delta^*)\right\} \\
        \end{aligned}
        \end{equation}
        As $\Sigma \vw \vw^T (\Sigma \vw) = \Sigma \vw (\vw^T \Sigma \vw)$, $\vq$ is an eigenvector of $\Sigma \vw \vw^T$. Therefore, $\exists \eta \in \sR$ s.t. $\Sigma \vw = \eta \vq$. When $N=2$, we have
        \begin{equation}
            \begin{aligned}
                \eta\Sigma\vw^* =& \vq\\
                =& \alpha\cdot{\exp\big(-\frac{((\vmu_{+}-\epsilon\Delta^*)^T\vw^*)^2}{2(\vw^*)^T\Sigma\vw^*}\big)}(\vmu_{+} - \epsilon\Delta^*) + (1-\alpha)\cdot{ \exp\big(-\frac{((\vmu_{-}-\epsilon\Delta^*)^T\vw^*)^2}{2(\vw^*)^T\Sigma\vw^*}\big)}(\vmu_{-} - \epsilon\Delta^*)\\
            \end{aligned}
        \end{equation}
        \begin{equation}
        \begin{aligned}
            &\eta
            \begin{bmatrix}
                \Sigma_1\vw^*_1\\
                \Sigma_2\vw^*_2
            \end{bmatrix}
            =\zeta\gamma_+
            \begin{bmatrix}
                \vmu_1 - \epsilon\vdelta_1\\
                \vmu_2 - \epsilon\vdelta_2
            \end{bmatrix}
            +(1-\zeta)\gamma_-
            \begin{bmatrix}
                \vmu_1 - \epsilon\vdelta_1\\
                -\vmu_2 - \epsilon\vdelta_2
            \end{bmatrix}
            \\
            \Longleftrightarrow\quad& 
            \eta
            \begin{bmatrix}
                \vw^*_1\\
                \vw^*_2
            \end{bmatrix}
            =\zeta\gamma_+
            \begin{bmatrix}
                \Sigma_1^{-1}(\vmu_1 - \epsilon\vdelta_1)\\
                \Sigma_2^{-1}(\vmu_2 - \epsilon\vdelta_2)
            \end{bmatrix}
            +(1-\zeta)\gamma_-
            \begin{bmatrix}
                \Sigma_1^{-1}(\vmu_1 - \epsilon\vdelta_1)\\
                \Sigma_2^{-1}(-\vmu_2 - \epsilon\vdelta_2)
            \end{bmatrix}\\
            \Longleftrightarrow\quad& 
            \eta
            \begin{bmatrix}
                \vw^*_1\\
                \vw^*_2
            \end{bmatrix}
            =\zeta\gamma_+
            \begin{bmatrix}
                \Sigma_1^{-1}\vmu_1\\
                \Sigma_2^{-1}\vmu_2
            \end{bmatrix}
            +(1-\zeta)\gamma_-
            \begin{bmatrix}
                \Sigma_1^{-1}\vmu_1\\
                -\Sigma_2^{-1}\vmu_2
            \end{bmatrix}
            -\epsilon\big(\zeta\gamma_++(1-\zeta)\gamma_-\big)
            \begin{bmatrix}
                \Sigma_1^{-1}\delta_1\\
                \Sigma_2^{-1}\delta_2
            \end{bmatrix}
        \end{aligned}
        \end{equation}
        with $\gamma_\pm$ as
        \begin{equation}
            \begin{aligned}
                \gamma_+ = & \exp\big(-\frac{((\vmu_{+}-\epsilon\Delta^*)^T\vw^*)^2}{2(\vw^*)^T\Sigma\vw^*}\big) \\
                \gamma_- = & \exp\big(-\frac{((\vmu_{-}-\epsilon\Delta^*)^T\vw^*)^2}{2(\vw^*)^T\Sigma\vw^*}\big) \\
            \end{aligned}
        \end{equation}
        and $\vmu_\pm$ as
        \begin{equation}
            \begin{aligned}
                \vmu_+ =
                \begin{bmatrix}
                    \vmu_1\\
                    \vmu_2
                \end{bmatrix},
                \vmu_- =
                \begin{bmatrix}
                    \vmu_1\\
                    -\vmu_2
                \end{bmatrix}
            \end{aligned}
        \end{equation}
        which denotes the conditioned mean of subgroups when the spurious feature aligns  $a_2=a_1$ or misaligns $a_2\neq a_1$ with the invariant feature.

        Considering $c_1, c_2$ as
        \begin{equation}
            \left\{
            \begin{aligned}
                c_1 =& \zeta\gamma_+ + (1-\zeta)\gamma_-\\
                c_2 =& \zeta\gamma_+ - (1-\zeta)\gamma_-
            \end{aligned}
            \right.
        \end{equation}
        we have
        \begin{equation}
            \begin{aligned}
            \eta
            \begin{bmatrix}
                \vw^*_1\\
                \vw^*_2
            \end{bmatrix}
            = 
            \begin{bmatrix}
                c_1\Sigma_1^{-1}\vmu_1\\
                c_2\Sigma_2^{-1}\vmu_2
            \end{bmatrix}
            -\epsilon c_1
            \begin{bmatrix}
                \Sigma_1^{-1}\delta_1\\
                \Sigma_2^{-1}\delta_2
            \end{bmatrix}
            \end{aligned}
        \end{equation}
        
    \end{proof}

\subsection{Proof of Theorem 6}
    \paragraph{Theorem 6 (Bayes-optimal linear classifier under perturbed data).} 
    \textit{The coefficient $c$ of the Bayes-optimal linear classifier under adversarial perturbations satisfies the equation
    \begin{align}
        c = \tanh(\frac{\log(\zeta) - \log(1-\zeta)}{2}-\phi(c))
    \end{align}
    where
    \[
    \phi(c)
    =
    \frac{(m_1 - 2\epsilon n_1 - c\epsilon n_2 + \epsilon^2 n_3)
          (\tau m_2 - \epsilon n_2)}
         {m_1 + \tau^2 m_2 - 2\epsilon n_1 - 2c\epsilon n_2 + \epsilon^2 n_3},
    \]
    and
    \[
    n_1 = \vmu_1^T \Sigma_1^{-1}\vdelta_1,\quad
    n_2 = \vmu_2^T \Sigma_2^{-1}\vdelta_2,\quad
    n_3 = (\Delta^*)^T \Sigma^{-1}\Delta^*.
    \]
    }

    \begin{proof}
        We have the Bayes-optimal classifier $\vw^*$ with $c_1, c_2$ as
        \begin{equation}
            \vw^*
            =
            [c_1\Sigma_1^{-1}(\vmu_1 - \epsilon\vdelta_1),\;
             \Sigma_2^{-1}(c_2\vmu_2 - \epsilon c_1 \vdelta_2)],
        \end{equation}
        \begin{equation}
            \left\{
            \begin{aligned}
                c_1 =& \zeta\gamma_+ + (1-\zeta)\gamma_-\\
                c_2 =& \zeta\gamma_+ - (1-\zeta)\gamma_-
            \end{aligned}
            \right.
        \end{equation}
        where $\gamma_\pm$ can be considered as functions of $c_1, c_2$ as follows:
        \begin{equation}
            \begin{aligned}
                \gamma_{+}(c_1, c_2) =& \exp\Big(-\frac{((\vmu_{+}-\epsilon\Delta^*)^T\vw^*)^2}{2(\vw^*)^T\Sigma\vw^*}\Big)\\
                =& \exp\Big(
                -\frac{1}{2}\frac{\big((\vmu_1 - \epsilon\vdelta_1)^Tc_1\Sigma_1^{-1}(\vmu_1 - \epsilon\vdelta_1) + (\vmu_2 - \epsilon\vdelta_2)^T\Sigma_2^{-1}(c_2\vmu_2 - \epsilon c_1\vdelta_2)\big)^2}
                {c_1^2(\vmu_1 - \epsilon\vdelta_1)^T\Sigma_1^{-1}(\vmu_1 - \epsilon\vdelta_1) + (c_2\vmu_2 - \epsilon c_1\vdelta_2)^T\Sigma_2^{-1}(c_2\vmu_2 - \epsilon c_1\vdelta_2)}
                \Big)\\
                =& \exp\Big(
                -\frac{1}{2}\frac{\big((\vmu_1 - \epsilon\vdelta_1)^T\Sigma_1^{-1}(\vmu_1 - \epsilon\vdelta_1) + (\vmu_2 - \epsilon\vdelta_2)^T\Sigma_2^{-1}(\frac{c_2}{c_1}\vmu_2 - \epsilon \vdelta_2)\big)^2}
                {(\vmu_1 - \epsilon\vdelta_1)^T\Sigma_1^{-1}(\vmu_1 - \epsilon\vdelta_1) + (\frac{c_2}{c_1}\vmu_2 - \epsilon \vdelta_2)^T\Sigma_2^{-1}(\frac{c_2}{c_1}\vmu_2 - \epsilon \vdelta_2)}\Big) := \gamma_+(c)
            \end{aligned}
        \end{equation}
        and
        \begin{equation}
            \begin{aligned}
                \gamma_{-}(c_1, c_2) =& \exp\Big(-\frac{((\vmu_{+}-\epsilon\Delta^*)^T\vw^*)^2}{2(\vw^*)^T\Sigma\vw^*}\Big)\\
                =& \exp\Big(
                -\frac{1}{2}\frac{\big((\vmu_1 - \epsilon\vdelta_1)^Tc_1\Sigma_1^{-1}(\vmu_1 - \epsilon\vdelta_1) + (-\vmu_2 - \epsilon\vdelta_2)^T\Sigma_2^{-1}(c_2\vmu_2 - \epsilon c_1\vdelta_2)\big)^2}
                {c_1^2(\vmu_1 - \epsilon\vdelta_1)^T\Sigma_1^{-1}(\vmu_1 - \epsilon\vdelta_1) + (c_2\vmu_2 - \epsilon c_1\vdelta_2)^T\Sigma_2^{-1}(c_2\vmu_2 - \epsilon c_1\vdelta_2)}
                \Big)\\
                =& \exp\Big(
                -\frac{1}{2}\frac{\big((\vmu_1 - \epsilon\vdelta_1)^T\Sigma_1^{-1}(\vmu_1 - \epsilon\vdelta_1) + (-\vmu_2 - \epsilon\vdelta_2)^T\Sigma_2^{-1}(\frac{c_2}{c_1}\vmu_2 - \epsilon \vdelta_2)\big)^2}
                {(\vmu_1 - \epsilon\vdelta_1)^T\Sigma_1^{-1}(\vmu_1 - \epsilon\vdelta_1) + (\frac{c_2}{c_1}\vmu_2 - \epsilon \vdelta_2)^T\Sigma_2^{-1}(\frac{c_2}{c_1}\vmu_2 - \epsilon \vdelta_2)}\Big) := \gamma_-(c)
            \end{aligned}
        \end{equation}

        Let $c = c_1/c_2$, we have
        \begin{equation}
            \begin{aligned}
                c =& \frac{c_2}{c_1} = \frac{\zeta\gamma_+(c) - (1-\zeta)\gamma_-(c)}{\zeta\gamma_+(c) + (1-\zeta)\gamma_-(c)} = \frac{\frac{\gamma_+(c)}{\gamma_-(c)} - \frac{(1-\zeta)}{\zeta}}{\frac{\gamma_+(c)}{\gamma_-(c)} + \frac{(1-\zeta)}{\zeta}}
            \end{aligned}
        \end{equation}
        Considering simplifying $\gamma_+(c) / \gamma_-(c)$, we have
        \begin{equation}
            \begin{aligned}
                \frac{\gamma_+(c)}{\gamma_-(c)} =& \exp\Big(\log(\gamma_+(c)) - \log(\gamma_-(c))\Big)\\
                =& \exp\Big(-\frac{(F+S_+)^2 - (F+S_-)^2}{D}\Big)\\
                =& \exp\Big(-\frac{(2F+S_++S_-)(S_+-S_-)}{D}\Big)
            \end{aligned}
        \end{equation}
        with $F = (\vmu_1 - \epsilon\vdelta_1)^T\Sigma_1^{-1}(\vmu_1 - \epsilon\vdelta_1), S_\pm = (\pm \vmu_2 - \epsilon\vdelta_2)^T\Sigma_2^{-1}(c\vmu_2 - \epsilon \vdelta_2), D = 2[(\vmu_1 - \epsilon\vdelta_1)^T\Sigma_1^{-1}(\vmu_1 - \epsilon\vdelta_1) + (c\vmu_2 - \epsilon \vdelta_2)^T\Sigma_2^{-1}(c\vmu_2 - \epsilon \vdelta_2)]$

        From the definitions of $D, F, S_\pm$, we have that
        \begin{equation}
        \begin{aligned}
            F + S_+ + S_- =& 2(\vmu_1 - \epsilon\vdelta_1)^T\Sigma_1^{-1}(\vmu_1 - \epsilon\vdelta_1) + (\vmu_2 - \epsilon\vdelta_2)^T\Sigma_2^{-1}(c\vmu_2 - \epsilon \vdelta_2) + (-\vmu_2 - \epsilon\vdelta_2)^T\Sigma_2^{-1}(c\vmu_2 - \epsilon \vdelta_2)\\
            =& 2(\vmu_1 - \epsilon\vdelta_1)^T\Sigma_1^{-1}(\vmu_1 - \epsilon\vdelta_1) - 2\epsilon\vdelta_2\Sigma_2^{-1}(c\vmu_2-\epsilon\vdelta_2) \\
            S_+ - S_- =& (\vmu_2 - \epsilon\vdelta_2)^T\Sigma_2^{-1}(c\vmu_2 - \epsilon \vdelta_2) - (-\vmu_2 - \epsilon\vdelta_2)^T\Sigma_2^{-1}(c\vmu_2 - \epsilon \vdelta_2)\\
            =& 2\vmu_2^T\Sigma_2^{-1}(c\vmu_2-\epsilon\vdelta_2)
        \end{aligned}
        \end{equation}
        
        As a result, the quotient is written as
        
        \begin{equation}
            \begin{aligned}
                \frac{\gamma_+(c)}{\gamma_-(c)} =& \exp\Big(-\frac{(2F+S_++S_-)(S_+-S_-)}{D}\Big)\\
                =& \exp\bigg(-\frac{\big(2(\vmu_1 - \epsilon\vdelta_1)^T\Sigma_1^{-1}(\vmu_1 - \epsilon\vdelta_1) - 2\epsilon\vdelta_2^T\Sigma_2^{-1}(c\vmu_2 - \epsilon \vdelta_2)\big)\big(2\vmu_2^T\Sigma_2^{-1}(c\vmu_2 - \epsilon \vdelta_2)\big)}{2[(\vmu_1 - \epsilon\vdelta_1)^T\Sigma_1^{-1}(\vmu_1 - \epsilon\vdelta_1) + (c\vmu_2 - \epsilon \vdelta_2)^T\Sigma_2^{-1}(c\vmu_2 - \epsilon \vdelta_2)]}\bigg)\\
                =& \exp\bigg(-\frac{2\big((\vmu_1 - \epsilon\vdelta_1)^T\Sigma_1^{-1}(\vmu_1 - \epsilon\vdelta_1) - \epsilon\vdelta_2^T\Sigma_2^{-1}(c\vmu_2 - \epsilon \vdelta_2)\big)\big(\vmu_2^T\Sigma_2^{-1}(c\vmu_2 - \epsilon \vdelta_2)\big)}{(\vmu_1 - \epsilon\vdelta_1)^T\Sigma_1^{-1}(\vmu_1 - \epsilon\vdelta_1) + (c\vmu_2 - \epsilon \vdelta_2)^T\Sigma_2^{-1}(c\vmu_2 - \epsilon \vdelta_2)}\bigg)\\
                =&:\exp(-2\phi(c))
            \end{aligned}
        \end{equation}
        with $\phi(c)$ as
        \begin{equation}
            \begin{aligned}
                \phi(c) =& \frac{\big((\vmu_1 - \epsilon\vdelta_1)^T\Sigma_1^{-1}(\vmu_1 - \epsilon\vdelta_1) - \epsilon\vdelta_2^T\Sigma_2^{-1}(c\vmu_2 - \epsilon \vdelta_2)\big)\big(\vmu_2^T\Sigma_2^{-1}(c\vmu_2 - \epsilon \vdelta_2)\big)}{(\vmu_1 - \epsilon\vdelta_1)^T\Sigma_1^{-1}(\vmu_1 - \epsilon\vdelta_1) + (c\vmu_2 - \epsilon \vdelta_2)^T\Sigma_2^{-1}(c\vmu_2 - \epsilon \vdelta_2)} \\
                =& \frac{(m_1 - 2\epsilon n_1 - c\epsilon n_2 + \epsilon^2 n_3) (\tau m_2 - \epsilon n_2)} {m_1 + \tau^2 m_2 - 2\epsilon n_1 - 2c\epsilon n_2 + \epsilon^2 n_3}
            \end{aligned}
        \end{equation}
        and $
        n_1 = \vmu_1^T \Sigma_1^{-1}\vdelta_1,\quad
        n_2 = \vmu_2^T \Sigma_2^{-1}\vdelta_2,\quad
        n_3 = (\Delta^*)^T \Sigma^{-1}\Delta^*.$
        Therefore, we have 
        \begin{equation}
            \begin{aligned}
                c =& \frac{\frac{\gamma_+(c)}{\gamma_-(c)} - \frac{(1-\zeta)}{\zeta}}{\frac{\gamma_+(c)}{\gamma_-(c)} + \frac{(1-\zeta)}{\zeta}} \\
                =& \frac{\exp(-2\phi(c)) -\frac{1-\zeta}{\zeta}}{\exp(-2\phi(c)) + \frac{1-\zeta}{\zeta}}\\
                =& \frac{\exp(-2\phi(c)) -\exp(\log(1-\zeta) - \log(\zeta))}{\exp(-2\phi(c)) + \exp(\log(1-\zeta) - \log(\zeta))}\\
                =& \tanh(\frac{-2\phi(c) - (\log(1-\zeta) - \log(\zeta))}{2})\\
                =& \tanh(-\phi(c) + \frac{\log(\zeta) - \log(1-\zeta)}{2})
            \end{aligned}
        \end{equation}
    \end{proof}

\subsection{Proof of Corollary 7}
    \paragraph{Corollary 7 (Group accuracy under perturbed data).}
    \textit{For the two-attribute setting, the accuracy of the Bayes-optimal linear classifier $\vw^*$ on subgroup $\va$ is
    \begin{align}
        \begin{aligned}
            & Acc_{\va}(\vw^*) = \frac{1}{2} + \operatorname{erf}\big(\frac{m_1 + a_1 a_2 c m_2 - \epsilon n_1 - \epsilon a_1 a_2 n_2}{\sqrt{2m_1 + c^2 m_2 - 2\epsilon n_1 - 2c\epsilon n_2 + \epsilon^2 n_3}}\big)
        \end{aligned}
    \end{align}
    }

    \begin{proof}
        We have the general form of group accuracy from Lemma 4, and we evaluate the group accuracy on clean test data, with the Bayes-optimal classifier $\vw^*$ fitted on perturbed data.
        \begin{equation}
        \begin{aligned}
            Acc_\va(\vw^*) =& \frac{1}{2} \Big( 1 + \operatorname{erf}\big(\frac{\vmu_\va^T\vw^*}{\sqrt{2(\vw^*)^T\Sigma^{-1}\vw^*}}\big) \Big)
        \end{aligned}
        \end{equation}
        With the Bayes-optimal classifier from Lemma 6, we have
        \begin{equation}
        \begin{aligned}
            (\vw^*)^T\Sigma^{-1}\vw^* =& (\vmu_1 - \epsilon\vdelta_1)^T\Sigma_1^{-1}(\vmu_1 - \epsilon\vdelta_1) + (c\vmu_2 - \epsilon \vdelta_2)^T\Sigma_2^{-1}(c\vmu_2 - \epsilon \vdelta_2) \\
            =& \vmu_1^T \Sigma_1^{-1}\vmu_1 - 2\epsilon \vdelta_1^T\Sigma_1^{-1}\vmu_1 + \epsilon^2 \vdelta_1^T\Sigma_1^{-1}\vdelta_1 + c^2\vmu_2^T \Sigma_2^{-1}\vmu_2 - 2c\epsilon \vdelta_2^T\Sigma_2^{-1}\vmu_2 + \epsilon^2 \vdelta_2^T\Sigma_2^{-1}\vdelta_2\\
            =& m_1 - 2\epsilon n_1 + c^2 m_2 - 2c\epsilon n_2 + \epsilon^2 n_3 \\
            \vmu_\va ^T \vw^* =& \vmu_1^T \Sigma_1^{-1}(\vmu_1 - \epsilon \vdelta_1) + a_1 a_2 \vmu_2^T \Sigma_2^{-1}(c\vmu_2 - \epsilon \vdelta_2) \\
            =&
            m_1 + c a_1 a_2 m_2
            - \epsilon n_1
            - \epsilon a_1 a_2 n_2
        \end{aligned}
        \end{equation}
        Therefore, the group accuracy becomes
        \begin{equation}
        \begin{aligned}
            Acc_{\va}(\vw^*) = \frac{1}{2} + \operatorname{erf}\big(\frac{m_1 + a_1 a_2 c m_2 - \epsilon n_1 - \epsilon a_1 a_2 n_2}{\sqrt{2m_1 + c^2 m_2 - 2\epsilon n_1 - 2c\epsilon n_2 + \epsilon^2 n_3}}\big)
        \end{aligned}
        \end{equation}
    \end{proof}

\subsection{Proof of Corollary 8}
    \paragraph{Corollary 8 (Consistency with the clean-data analysis).}
    \textit{Setting $\epsilon = 0$ in the derived Bayes-optimal classifier and group accuracy recovers the clean-data Bayes-optimal classifier and group accuracy characterized in Section 3.4.
    }
    \begin{proof}
        When $\epsilon=0$, the Bayes-optimal classifier becomes
        \begin{equation}
        \begin{aligned}
            \vw^*
            = &
            [\Sigma_1^{-1}(c_1\vmu_1 - \epsilon\vdelta_1),\;
             \Sigma_2^{-1}(c_2\vmu_2 - \epsilon\vdelta_2)] \\
            = &
            [c_1\Sigma_1^{-1}\vmu_1,\;
             c_2\Sigma_2^{-1}\vmu_2]
        \end{aligned}
        \end{equation}
        where $c = c_2 / c_1$ equals the solution of the Bayes-optimal classifier $\tau$ on clean data.
        As $\epsilon=0$ sets $n_1=n_2=n_3=0$. Then the group accuracy becomes
        \begin{equation}
            \begin{aligned}
                Acc_{\va}(\vw^*; \epsilon \Delta^*) =& \frac{1}{2} \Big(1 + \operatorname{erf} \big(\frac{m_1 + a_1 a_2 c m_2 - 2 \epsilon n_1 - \epsilon(c + a_1 a_2) n_2 + \epsilon^2 n_3,}{\sqrt{2m_1 + c^2 m_2 - 2\epsilon n_1 - 2c\epsilon n_2 + \epsilon^2 n_3}} \big) \Big) \\
                = & \frac{1}{2} \Big(1 + \operatorname{erf} \big(\frac{m_1 + a_1 a_2 c m_2}{\sqrt{2\left(m_1 + c^2 m_2\right)}} \big) \Big) \\
                = & \frac{1}{2} \Big(1 + \operatorname{erf} \big(\frac{m_1 + a_1 a_2 \tau m_2}{\sqrt{2\left(m_1 + \tau^2 m_2\right)}} \big) \Big)
            \end{aligned}
        \end{equation}
        which also recovers the group accuracy on clean data in Corollary 3.
    
    \end{proof}

\end{document}